\documentclass{ifacconf}

\usepackage[pdftex]{graphicx}
\usepackage{natbib}

\usepackage[cmex10]{amsmath}
\usepackage{amssymb}
\usepackage{import}
\usepackage{bm}
\usepackage{color}
\usepackage{units}
\usepackage{url}
\usepackage{float}
\usepackage{epstopdf}
\usepackage{stmaryrd}
\usepackage{units}
\usepackage{siunitx}
\usepackage{outlines}
\usepackage{boldline}
\usepackage[flushleft]{threeparttable}

\usepackage{mathtools}
\usepackage{color,soul}

\usepackage{algorithm}
\usepackage{algorithmicx}
\usepackage{algpseudocode}
\algrenewcommand\algorithmicrequire{\textbf{Input:}}
\algrenewcommand\algorithmicensure{\textbf{Output:}}

\newcommand{\mnorm}[1]{\left\| #1 \right\|}

\newcommand{\mvec}[1]{\boldsymbol{\mathbf{#1}}}

\newcommand{\Tau}{\mathcal{T}}

\newcommand{\mnoshow}[1]{}

\newcommand{\figref}[1]{Fig.~\ref{#1}}
\newcommand{\secref}[1]{Section~\ref{#1}}

\newcommand*{\transpose}{\top}

\newcommand{\tr}{\mathrm{trace}}

\DeclareMathOperator*{\argmin}{arg\,min}
\DeclareMathOperator*{\argmax}{arg\,max}

\newtheorem{lemma}{Lemma}

\begin{document}
\begin{frontmatter}

\title{Robust Feature-Based Point Registration Using Directional Mixture Model
\thanksref{footnoteinfo}} 
 
\thanks[footnoteinfo]{This work was supported in part by the National Science Foundation Graduate Research Fellowship under Grant No. DGE 1106400.}

\author[First]{Saman Fahandezh-Saadi} 
\author[Second]{Di Wang} 
\author[First]{Masayoshi Tomizuka}

\address[First]{Mechanical Systems Control (MSC) Lab, University of California, Berkeley, USA. e-mail: \{samanfahandej, tomizuka\}@berkeley.edu}
\address[Second]{Visual Cognitive Computing and Intelligent Vehicle (VCC\&IV) Lab, Xi'an Jiaotong University, Xi’an, P.R. China. e-mail: de2wang@stu.xjtu.edu.cn}

\begin{abstract}                
This paper presents a robust probabilistic point registration method for estimating the rigid transformation (i.e. rotation matrix and translation vector) between two pointcloud dataset. The method improves the robustness of point registration and consequently the robot localization in the presence of outliers in the pointclouds which always occurs due to occlusion, dynamic objects, and sensor errors. The framework models the point registration task based on directional statistics on a unit sphere. In particular, a Kent distribution mixture model is adopted and the process of point registration has been carried out in the two phases of Expectation-Maximization algorithm. The proposed method has been evaluated on the pointcloud dataset from LiDAR sensors in an indoor environment.
\end{abstract}


\end{frontmatter}

\section{Introduction}\label{Intro}
The problem of point registration or matching plays a crucial role in many engineering applications and scientific disciplines from robot navigation, odometry and autonomous vehicles to graphics, object modeling, and medical imaging. In all of these applications, it is important to acquire a very accurate point registration, despite the sensors errors and limitations. The goal of point registration is to use 3D dataset observation and try to find the best rigid transformation hypothesis that maps one frame to the next. The rigid transformation $\left(\mvec{R}, \mvec{t}\right)\in SE(3)$ consists of a rotation matrix $\mvec{R} \in SO(3)$ and a translation vector $\mvec{t}\in \mathbb{R}^3$; and a 3D point $\mvec{x}\in \mathbb{R}^3$ can be accordingly transformed rigidly as $\mvec{T}(\mvec{x})=\mvec{R}\mvec{x}+\mvec{t}$. 

Recently, the point registration algorithms have received extensive focus from academia and industries in autonomous driving, since the rapidly developed 3D sensing technology endows the intelligent vehicle the capacity of accurate mapping and localization. Superiority of 3D LiDAR which can reliably employed during night, bad weather (e.g. rain, snow), and in terms of computing complexity, makes it a suitable choice of 3D perception for intelligent vehicles. However, the limited sensor's coverage, intrinsic sensor errors, and outliers caused by occlusion or unpredictable traffic participators in the real traffic scenarios are formidable challenges which require to be addressed when trying to deploy the point registration algorithm in intelligent vehicles.

\begin{figure}[!t]
    \centering
    \includegraphics[width=0.475\linewidth]{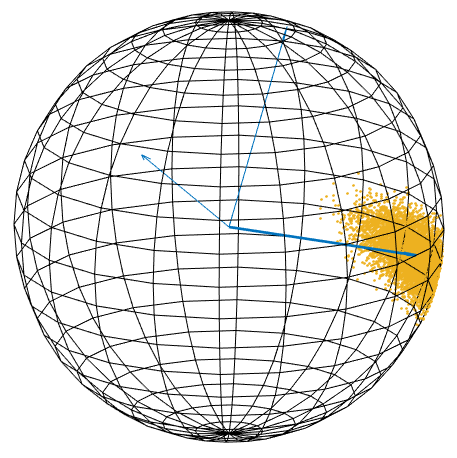}
    \includegraphics[width=0.475\linewidth]{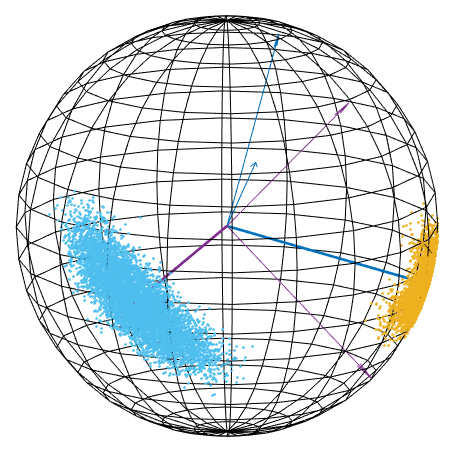}
    \caption{Two data samples drawn from Kent distribution. \textbf{Left:} Samples drawn from a single Kent PDF. Arrows show $\mvec{\mu}$ the mean direction of samples as well as $\mvec{\gamma}_1, \mvec{\gamma}_2$ major and minor axes which depend on the orientation of samples. \textbf{Right:} Shows a mixture of two Kent distributions with different parameters. The pointcloud surface normals with different concentrations, can be modeled as a mixture of Kent distributions.}
    \label{fig:samples}
\end{figure}
A standard approach for point registration problem is the Iterative Closest Point (ICP) algorithm (\cite{ICP}) which solves the problem in two iterative steps: finding the correspondence points and then minimizing the sum of squared residuals to find the best transformation. The ICP method is sensitive to the outliers and the solution is heavily influenced by the initialization. There are several variants of ICP method that try to improve the original method in different regards: ICPp2pl (\cite{p_to_pl_ICP}) ICPpl2pl (\cite{pl_to_pl_ICP}), GICP (\cite{GICP}). In all of these methods both rotation matrix and translation vector are found together iteratively. In addition, the correspondence phase is hard-assignment or one-to-one.

Our proposed method solves the problem of point registration in a probabilistic fashion by incorporating the orientational information from the pointcloud. The method is mainly motivated by the fact that it can robustify the registration to noise and outliers by incorporating surface normals. In particular, the surface normals are assumed to be drawn from a directional distribution called Kent distribution. Directional statistic is a suitable way to represent pointcloud features like surface normals, since they represent a directional pattern in the pointcloud and also they are invariant with respect to translation. In this approach the estimation of rigid transformation can be decoupled into two steps: first, the rotation matrix between two frames can be estimated using the surface normals, and second, the translation vector can be found using the positional information of the pointcloud. The decoupling of rotation matrix from translation vector is also addressed in \cite{Delio}.

Many previous works use the probabilistic approach (i.e. soft-assignment) for the correspondence phase (\cite{soft_assgin1, soft_assgin2, soft_assign3_GMM}). In contrast to hard-assignment or one-to-one correspondence in classical ICP methods, in this method the points are associated to each other in two frames according to a probability distribution. \cite{soft_assign3_GMM} formulated the point registration problem as a maximum likelihood (ML) estimation problem by using Gaussian mixture model (GMM). \cite{multiple_GMM} also use GMM to describe and register multiple pointclouds jointly that are assumed to be drawn from an underlying distribution. There are other variants of GMM method with different noise and outlier conditions (\cite{robust_gmm, rigid_gmm}).

Recently, \cite{RG_register} and \cite{GIMLOP} proposed point registration with Von Mises–Fisher distribution which is also a directional distribution defined on a unit sphere for 3D points. The method is a hybrid method that combines both directional and positional information into a hybrid of Gaussian and Von Mises distributions. Although the method improves the registration with regard to outliers, its isotropic assumption is not realistic in the real world applications. In contrast, Kent distribution accounts for anisotropic (i.e. the ovalness of surface normals on a unit sphere) dataset as seen in \figref{fig:samples}. This will further address the problem of outliers in dataset. Based on the proposed statistical framework, the iterations of finding correspondences and updating transformations are now considered as a type of Expectation-Maximization (EM) procedure.

The rest of this paper is organized as follows; first in \secref{method} we start by preliminaries and introducing nomenclatures, including the definition of Kent distribution. Then we introduce the mixture model that describes the probabilistic relationship between points in two frames. In \secref{algorithm} we detail the steps in the EM algorithm in the proposed method. In \secref{Expriment} we present the results from the experiments on ETH dataset, and finally in \secref{conclusion} we make the concluding remarks and discuss the future directions. 

\section{Methods}
\label{method}
\subsection{Preliminaries} 
Throughout the paper we use the following notations for pointclouds:
\begin{itemize}
    \item $M$, $N$-- number of points in the \emph{model} and \emph{observed} pointclouds, respectively, 
    \item $\mathbf{Y}=\{\mvec{y}_i\}_{m= 1,
    \hdots, M},\ \mvec{y}_i\in \mathbb{R}^3$-- the \emph{model} pointcloud,
    \item $\mathbf{X}=\{\mvec{x}_i\}_{n= 1,
    \hdots, N},\ \mvec{x}_i\in \mathbb{R}^3$-- the \emph{observed} pointcloud,
    \item $\mathbf{\tilde{Y}}=\{\mvec{\tilde{y}}_m\in \mathbb{R}^3:\mnorm{\mvec{\tilde{y}}_i}_2=1\}_{i= 1,
    \hdots, M}$-- the surface normals of \emph{model} data,
    \item $\mathbf{\tilde{X}}=\{\mvec{\tilde{x}}_n\in \mathbb{R}^3:\mnorm{\mvec{\tilde{x}_i}}_2=1\}_{i= 1,
    \hdots, N}$-- the surface normals of \emph{observed} data,
    \item $\mvec{\mu}_y$ the positional mean of the \emph{model} pointcloud,
    \item $\mvec{\mu}_x$ the positional mean of the \emph{observed} pointcloud, 
    \item $\mvec{z}_n$, $1 \leq n\leq N$, the hidden random variable in the mixture model. 
\end{itemize}
\begin{figure}[!t]
\begin{minipage}[c]{.5\textwidth}
    \centering
    \includegraphics[width=0.4\textwidth]{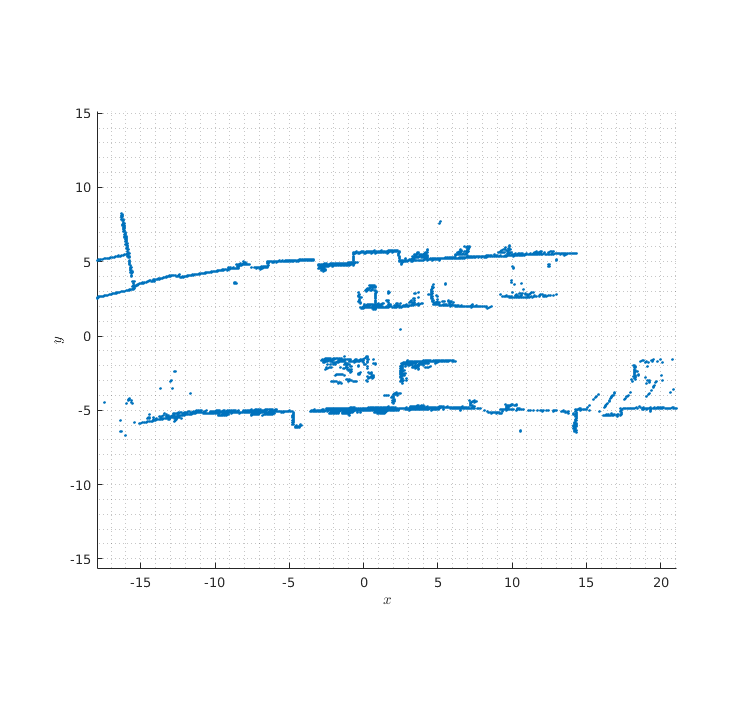}
    \includegraphics[width=0.4\textwidth]{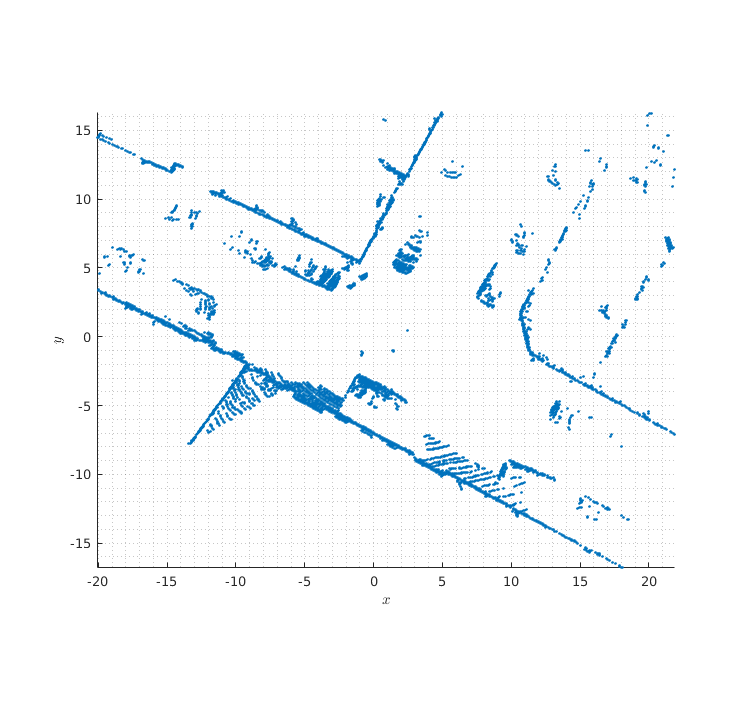}
\end{minipage}
\begin{minipage}[c]{.5\textwidth}
    \centering
    \includegraphics[width=0.45\textwidth]{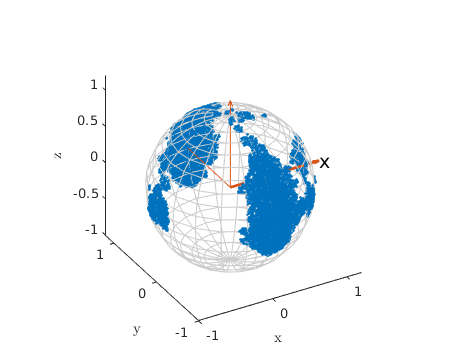}
    \includegraphics[width=0.45\textwidth]{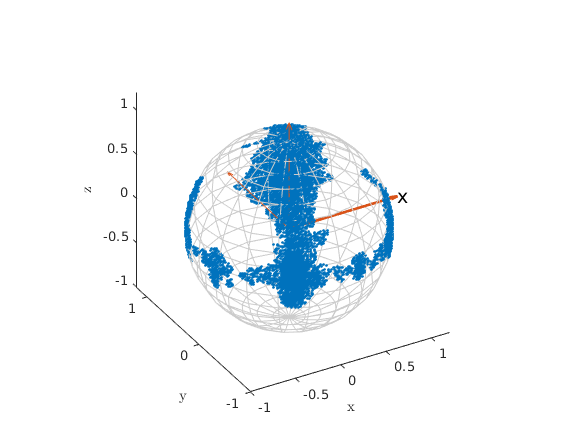}
\end{minipage}
    \caption{Example from the KITTI dataset \cite{KITTI}. Figures show two urban scenes pointcloud data and associated surface normals on a unit sphere. As seen, as the vehicle moves, the surface normals are also moving on the unit sphere which indicates the rotation between consecutive frames.}
    \label{fig:tw_scenes}
\end{figure}
Kent distribution is the spherical analogous of the bivariate Gaussian distribution which was introduced by \cite{Kent1982}. Since the distribution is defined over the set of unit vector on a sphere, it is suitable for representing surface normals of data points in a pointcloud. The distribution can be represented with 5 parameters ($\kappa, \beta, \mvec{\mu}, \mvec{\gamma}_1, \mvec{\gamma}_2$) as\footnote{In the original paper the distribution was called Fisher-Bingham with $5$ parameters, therefore we use the same notation $\text{FB}_5(.)$ in this paper as well.},
\begin{align*}
    \text{FB}_5(\mvec{\tilde{x}}) = c(\kappa, \beta)^{-1} \exp \big\{\kappa \mvec{\mu}^\transpose\mvec{\tilde{x}} + \beta \big[(\mvec{\gamma}_{1}^\transpose \mvec{\tilde{x}})^2-(\mvec{\gamma}_{2}^\transpose \mvec{\tilde{x}})^2\big]\big\},
\end{align*}
where $\kappa\geq 0$ is the \emph{concentration}, $\beta\geq 0$ describes \emph{ovalness}, $\mvec{\mu}\in \mathbb{R}^3$ is the \emph{mean direction}, and $\mvec{\gamma}_1, \mvec{\gamma}_2 \in \mathbb{R}^3$ are the \emph{major axis} and \emph{minor axis} of the orientation of the distribution, respectively. Together, $\mvec{\Gamma} = (\mvec{\mu}, \mvec{\gamma}_1, \mvec{\gamma}_2)$ creates an orthogonal matrix representing the total orientation of data on the sphere. $c(\kappa, \beta)$ is the normalization term which can be expressed, in general, with an infinite sequence of the Modified Bessel functions, but it can be simplified under the assumption of $2\beta<\kappa$ and large $\kappa$ as $c(\kappa, \beta) = \frac{2\pi e^\kappa}{(\kappa^2-4\beta^2)^{1/2}}$.
\subsection{Problem Formulation}
The point registration problem can be modeled as a mixture of Kent distributions. The distribution which describes the probability of an observed data point corresponding to a model data point has the form (\cite{GIMLOP}),
\begin{align}\label{eq:fb5}
    &\text{FB}_5(\mvec{\tilde{x}}_n| \mvec{z}_n=\mvec{\tilde{y}}_m) = \nonumber\\
    & c(\kappa, \beta)^{-1}\exp \big\{\kappa \mvec{\tilde{y}}_m^\transpose\mvec{\tilde{x}}_n + \beta \big[(\mvec{\gamma}_{1}^\transpose \mvec{\tilde{x}}_n)^2-(\mvec{\gamma}_{2}^\transpose \mvec{\tilde{x}}_n)^2\big]\big\},
\end{align}
where $\mvec{\tilde{y}}_m$ are substituted as the \emph{mean directions}, and consequently the marginal distribution of an observed data point $\mvec{\tilde{x}}_n$ can be computed as,
\begin{align}\label{eq:margin}
    p(\mvec{\tilde{x}}_n) = \sum_{m=1}^{M} \pi_m \text{FB}_5(\mvec{\tilde{x}}_n | \mvec{z}_n=\mvec{\tilde{y}}_m) + \pi_0 p_0(\mvec{\tilde{x}}_n).
\end{align}
In general, the membership probabilities are part of parameter estimation in mixture models, but here they are assumed to be constant and identical. We also add a uniform distribution which accounts for the noise/outlier in the data points as $p_0(\mvec{\tilde{x}}_n)=\frac{1}{N}$. $\pi_0$ can be chosen empirically based on the approximate amount of noise/outlier in the dataset. \figref{fig:tw_scenes} shows an example of two urban scenes and the estimated surface normals which represents a mixture of Kent distributions. Normals rotate on the sphere in consecutive frames and are invariant to the translation along the trajectory of the vehicle. 

By assuming that all surface normals of the observed data points $\mvec{\tilde{x}}_1, \hdots, \mvec{\tilde{x}}_N$ are independent, the likelihood function is
\begin{align}\label{eq:likelihood}
    L(\mvec{\Theta}) = p(\mvec{\tilde{X}} | \mvec{\tilde{Y}}; \mvec{\Theta}) = \prod_{n=1}^{N} p(\mvec{\tilde{x}}_n),
\end{align}
and the log-likelihood function can be computed as
\begin{align}\label{eq:ll}
    \log L(\mvec{\Theta}) &= \sum_{n=1}^{N} \log p(\mvec{\tilde{x}}_n) \\
    &= \sum_{n=1}^{N} \log \sum_{m=1}^{M} \left( \pi_m \text{FB}_5(\mvec{\tilde{x}}_n|\mvec{z}_n=\mvec{\tilde{y}}_m) +\pi_0 p_0(\mvec{\tilde{x}}_n) \right), \nonumber
\end{align}
where the parameters are $\mvec{\Theta} = (\kappa, \beta, \mvec{\gamma}_1, \mvec{\gamma}_2, \mvec{R})$. The matrix $\mvec{R}$ rotates the surface normals of the \emph{observed} pointcloud $\mvec{\tilde{X}}$ to the surface normals of the \mvec{model} pointcloud $\mvec{\tilde{Y}}$, and has been expressed implicitly in \eqref{eq:fb5} as $\mvec{\tilde{y}}_m = \mvec{R}\mvec{\tilde{x}}_n$. Rotating surface normals does not change other parameters in the distribution as stated in the following lemma.
\begin{lemma} If \mvec{R} is an orthogonal matrix, and $\mvec{\tilde{x}} \sim \text{FB}_5(\kappa, \beta, \mvec{\mu}, \mvec{\gamma}_1, \mvec{\gamma}_2)$, then, $\mvec{\tilde{y}} \sim \text{FB}_5(\kappa, \beta, \mvec{R}\mvec{\mu}, \mvec{R}\mvec{\gamma}_1, \mvec{R}\mvec{\gamma}_2)$ when $\mvec{\tilde{y}} = \mvec{R}\mvec{\tilde{x}}$.
\end{lemma}
\begin{pf}
From the definition of Kent distribution we know that we always have $\mvec{\tilde{x}}^* \sim \text{FB}_5(\kappa, \beta, \mvec{I})$ when $\mvec{\tilde{x}}^* = \mvec{\Gamma}^\transpose \mvec{\tilde{x}}$. Then if $\mvec{\tilde{x}}^* = (\mvec{R}\mvec{\Gamma})^\transpose \mvec{\tilde{y}}$, therefore, $\mvec{\tilde{x}}^* = \mvec{\Gamma}^\transpose \mvec{R}^\transpose \mvec{\tilde{y}}$, which means $\mvec{\tilde{y}} = \mvec{R}\mvec{\tilde{x}}$. 
\end{pf}
This is also intuitive that the orientation parameter does not change the compactness of the Kent distribution, since the rotation matrix is orthogonal, therefore $\kappa$ and $\beta$ are preserved. This helps us to estimate the change of the orientation of surface normals without considering any changes in the compactness and ovalness.

Log-likelihood function in \eqref{eq:ll} assumes a prior knowledge of point correspondence (i.e. the correspondence between points in two pointclouds). However, in practice we don't know those correspondences and we refer to the observed data as \emph{incomplete}. Since the correspondence between surface normals in \emph{observed} and \emph{model} data is missing, the hidden variable $\mvec{z}_n$ is introduced in EM algorithm to represent their probabilistic assignment. We describe an EM-like approach in the next section in order to iteratively solve for \emph{incomplete-data} version of the maximum likelihood (ML) problem.

\begin{algorithm}[t]
 \caption{Main Algorithm}
    \begin{algorithmic}[1]
        \Require observed and model pointclouds: $\mvec{X}$, $\mvec{Y}$.
        \Ensure  $\mvec{R}$, $\mvec{t}$
        \State Surface normals estimation $\mvec{\tilde{X}}$, $\mvec{\tilde{Y}}$ according to \secref{surf_nromals_sub_sec}
        \State Divide $\mvec{\tilde{Y}}$, $\mvec{\tilde{X}}$ to $K$ groups according to \secref{spherical_k_mean}
        \State Update $\mvec{\Theta}_k$ in parallel for each group $k=1,\hdots,K$ according to Algorithm \ref{alg:em_like_alg}.
        \State Average $\mvec{R}_k$ according to equation \eqref{eq:rot_avg}.
        \State Update $\mvec{t}$ according to equation \eqref{eq:translatino_eq}.
        \State Return $\mvec{R}$, $\mvec{t}$.
    \end{algorithmic}
 \label{alg:main_alg}
\end{algorithm}
\section{Point Registration Algorithm}
\label{algorithm}
In this section we describe the proposed approach for 3D point registration using Kent distribution. The method relies on decoupling of rotation from translation by estimating rotation based on surface normals and translation from the positional data points. Algorithms \ref{alg:main_alg} and \ref{alg:em_like_alg} outline the steps of the method.
\subsection{Surface Normals Estimation}\label{surf_nromals_sub_sec}
Several methods exist to estimate the surface normals associated with each point in the pointcloud (\cite{Compare_surface_normals}). In this paper we compute the normals using principal component analysis (PCA) method. To improve the performance of the proposed method, we pre-process the surface normals with: a) orientation consistency and b) outliers removal. The orientations of surface normals computed by PCA are ambiguous (i.e. it is impossible to solve for the sign of normal vectors). To ensure we have consistent surface normals in the algorithm, we use a defined viewpoint $\mvec{v}_p$ to make their orientation consistent (\cite{consistent_normals}). All normals $\mvec{n}_i$ of points $\mvec{x}_i$ with the same directions should satisfy $\mvec{n}_i\cdot(\mvec{v}_p-\mvec{x}_i) > 0$. We also use a modified version of outlier removal method for the surface normals. We utilize \emph{cosine similarity} in order to remove normals that are far away from their \textit{mean direction}. The method is detailed in \cite{outlier_removal} and \cite{outlier}.
\subsection{Clustering with Spherical $k$-means}\label{spherical_k_mean}
\begin{figure}[!t]
    \centering
    \includegraphics[width=\linewidth]{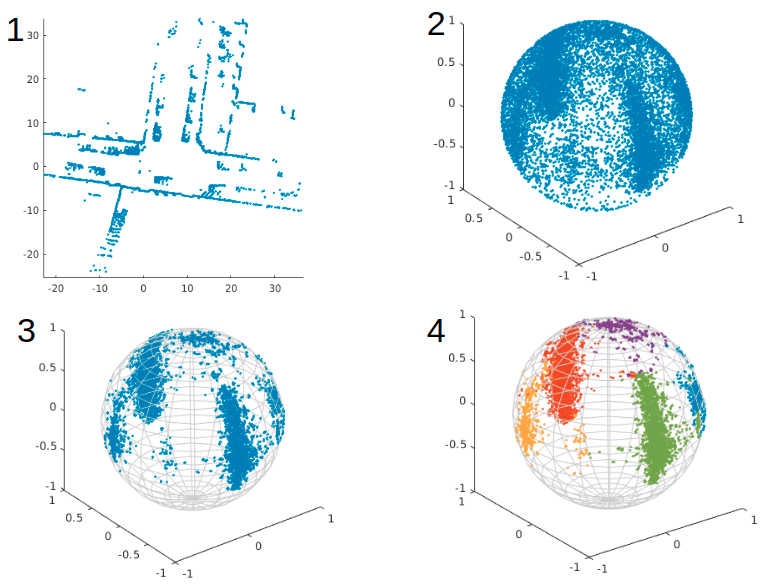}
    \caption{The process of preparing surface normals from pointcloud data. 1) pointcloud 2) estimating surface normals 3) outlier removal 4) \emph{spherical} clustering.}
    \label{fig:four_steps}
\end{figure}

In the most of urban scene pointclouds, the majority of surface normals are estimated based on prominent surfaces such as buildings which generate the pattern of surface normals concentrated in specific areas on the sphere (for example see \figref{fig:tw_scenes}). Therefore we use spherical $k$-means approach (\cite{spherical_k_means}) to cluster surface normals and use those normals from the same cluster in both observed and model data points. This helps to run the algorithm in parallel for the smaller size of data points in each cluster and finally take an average to find the overall rotation matrix in each step.  It is worth noting that the spherical clustering is closely related to ML estimate of Von-Mises distribution which is the same as Kent distribution when $\beta=0$ (\cite{k_means_von_mises}). \figref{fig:four_steps} depicts steps in order to prepare surface normals for the EM algorithm which will be explained in the following subsection.
\subsection{EM-like Algorithm}
A variant of EM approach is adopted here to estimate the parameter $\mvec{\Theta}$. As stated in \secref{method}, since the point correspondence is not available (i.e. the observed data is viewed as being \textit{incomplete}), log-likelihood function in \eqref{eq:ll} cannot be optimized directly. The \textit{complete-data log-likelihood} function is given by 
\begin{align}\label{eq:Qfunc_z} 
    \log L_c(\mvec{\Theta}) = \sum_{n=1}^{N} \sum_{m=1}^{M} &P(\mvec{z}_n|\mvec{\tilde{x}}_n) \bigg(\log \pi_0 + \log p_0(\mvec{\tilde{x}}_n)  \\
    &+\log \pi_m +\log \text{FB}_5(\mvec{\tilde{x}}_n|\mvec{z}_n=\mvec{\tilde{y}}_m)\bigg), \nonumber
\end{align}
where $P(\mvec{z}_n|\mvec{\tilde{x}}_n)$ is the posterior correspondence probability that links the observed data to the realization of hidden variables. The parameters will be found by minimizing the negative of the conditional expectation of $\log L_c(\mvec{\Theta})$ iteratively in the following two steps.

\begin{algorithm}[t]
 \caption{EM-like Algorithm}
     \begin{algorithmic}[1]
        \renewcommand*{\algorithmicrequire}{\textbf{Initialization:}}
        \Require $\kappa, \beta, \mvec{\gamma}_1, \mvec{\gamma}_2, \mvec{R}$
        \While{not converged}
            \State \textit{E-step:} Compute posterior probabilities $\tau_{mn}$ according to equation \eqref{eq:posterior}
            \State \textit{M-step:}
            \State Update $\mvec{R}$ according to equation \eqref{solve_for_R}
            \State Compute parameters using moment estimates according to \secref{moment_estimate}
        \EndWhile
        \State Return $\kappa, \beta, \mvec{\gamma}_1, \mvec{\gamma}_2, \mvec{R}$
     \end{algorithmic}
     \label{alg:em_like_alg}
\end{algorithm}

\subsection{E-step}
The hidden correspondence-label $\mvec{z}_n$ will be handled by the \textit{E-step}, which is computing the posterior probability distribution $\tau_{mn}$. This posterior distribution plays the role of \textit{soft-assignment} between the model normals $\mvec{\tilde{y}}_m$ and the observed normals $\mvec{\tilde{x}}_n$, and can be computed as
\begin{align}\label{eq:posterior}
    \tau_{mn}=P(\mvec{z}_n=\mvec{\tilde{y}}_m|\mvec{\tilde{x}}_n) = \frac{\pi_m \text{FB}_5(\mvec{\tilde{x}}_n |\mvec{z}_n=\mvec{\tilde{y}}_m)}{p(\mvec{\tilde{x}}_n)},
\end{align}
for $m=1,\hdots,M$, $n=1,\hdots,N$, and where the denominator is defined in \eqref{eq:margin}. The posterior probability of assigning observed normal to an outlier is $\tau_{0n}=1 - \sum_{m=1}^{M} \tau_{mn}$. By replacing $\tau_{mn}$ in \eqref{eq:Qfunc_z} we have the function that needs to be maximized in the \textit{M-step},
\begin{align}\label{eq:Qfunc}
    Q(\mvec{\Theta}) = \sum_{n=1}^{N} \sum_{m=1}^{M} &\tau_{mn} \big(\log \pi_0 + \log p_0(\mvec{\tilde{x}}_n)  \\
    &+\log \pi_m +\log \text{FB}_5(\mvec{\tilde{x}}_n|\mvec{z}_n=\mvec{\tilde{y}}_m)\big). \nonumber
\end{align}
\subsection{M-step}
This step requires to find the parameter $\mvec{\Theta}$ of the distribution by maximizing the $Q(\mvec{\Theta})$ function
\begin{align}\label{eq:max_Q}
    \mvec{\Theta}^* = \argmax_{\mvec{\Theta}}\ Q(\mvec{\Theta}).
\end{align}
Although the optimization problem in \eqref{eq:max_Q} has closed form solution for some types of probability distributions, unfortunately due to the orthogonality constraints on Kent distribution parameters and rotation matrix, this constrained nonlinear problem is hard to solve. Therefore we solve the problem in two sub-steps: finding rotation matrix and then updating Kent distribution parameters.
\subsubsection{Rotation Matrix Update:}
Substituting in \eqref{eq:Qfunc} we have
\begin{align}\label{eq:Q_expand}
    Q(\mvec{\Theta}) &= \sum_{m=1}^{M} \sum_{n=1}^{N} \tau_{mn} \bigg(\log \pi_0 + \log p_0(\mvec{\tilde{x}}_n) \nonumber \\
    &+ \log \pi_m + \log c(\kappa, \beta)^{-1} \nonumber \\ 
    &+\kappa\mvec{\tilde{y}}_m^\transpose\mvec{R}\mvec{\tilde{x}}_n + \beta \big[(\mvec{\gamma}_1^\transpose\mvec{R}\mvec{\tilde{x}}_n)^2-
     (\mvec{\gamma}_2^\transpose\mvec{R}\mvec{\tilde{x}}_n)^2\big]\bigg).
\end{align}
The optimal rotation matrix can be found by maximizing $Q$-function with respect to the rotation matrix which belongs to the set of \textit{special orthogonal} matrices $SO(3)$ defined as
\begin{align}\label{eq:constrained}
    SO(3) = \{\mvec{R} \in \mathbb{R}^{3\times3} | \mvec{R}^\transpose\mvec{R} = \mvec{I}, \det{\mvec{R}} = 1\}.
\end{align} 
After removing constant terms and also minimizing negative of $Q$-function, the constraint optimization problem can be written as,
\begin{align}\label{eq:argminR}
    &\mvec{R}^* = \argmin_{\mvec{R}\in SO(3)}\ -Q(\mvec{\Theta}) 
    = \argmin_{\mvec{R}\in SO(3)} \nonumber \\
    &\sum_{m=1}^{M} \sum_{n=1}^{N} \tau_{mn} \big(-\kappa\mvec{\tilde{y}}_m^\transpose\mvec{R}\mvec{\tilde{x}}_n + \beta \big[(\mvec{\gamma}_2^\transpose\mvec{R}\mvec{\tilde{x}}_n)^2-
     (\mvec{\gamma}_1^\transpose\mvec{R}\mvec{\tilde{x}}_n)^2\big]\big) \nonumber \\
    &= \argmin_{\mvec{R}\in SO(3)} -\kappa \sum_{m=1}^{M} \sum_{n=1}^{N} \tau_{mn}\mvec{\tilde{y}}_m^\transpose\mvec{R}\mvec{\tilde{x}}_n  \nonumber \\
    &+\beta \sum_{m=1}^{M} \sum_{n=1}^{N} \tau_{mn} (\mvec{\tilde{x}}_n^\transpose\mvec{R}^\transpose (\mvec{\gamma}_2\mvec{\gamma}_2^\transpose - \mvec{\gamma}_1\mvec{\gamma}_1^\transpose)\mvec{R}\mvec{\tilde{x}}_n).
\end{align}
Using matrix trace and its \textit{cyclic property}, the first term above can be rewritten as
\begin{align*}
    &\sum_{m=1}^{M} \sum_{n=1}^{N} \tau_{mn} \mvec{\tilde{y}}_m^\transpose \mvec{R} \mvec{\tilde{x}}_n
    = \sum_{m=1}^{M} \sum_{n=1}^{N} \tr(\tau_{mn}\mvec{\tilde{y}}_m^\transpose \mvec{R} \mvec{\tilde{x}}_n) =\\
    & \sum_{m=1}^{M} \sum_{n=1}^{N} \tr(\mvec{R} \tau_{mn}\mvec{\tilde{x}}_n\mvec{\tilde{y}}_m^\transpose) = \tr(\mvec{R}\sum_{m=1}^{M} \sum_{n=1}^{N} \tau_{mn} \mvec{\tilde{x}}_n \mvec{\tilde{y}}_m^\transpose).
\end{align*}
The second term in \eqref{eq:argminR} also can be simplified the same way,
\begin{align*}
    &\sum_{m=1}^{M} \sum_{n=1}^{N} \tau_{mn}\mvec{\tilde{x}}_n^\transpose\mvec{R}^\transpose (\mvec{\gamma}_2\mvec{\gamma}_2^\transpose - \mvec{\gamma}_1\mvec{\gamma}_2^\transpose)\mvec{R}\mvec{\tilde{x}}_n =\\
    &\sum_{m=1}^{M} \sum_{n=1}^{N} \tr\big(\tau_{mn}\mvec{\tilde{x}}_n^\transpose\mvec{R}^\transpose (\mvec{\gamma}_2\mvec{\gamma}_2^\transpose - \mvec{\gamma}_1\mvec{\gamma}_1^\transpose)\mvec{R}\mvec{\tilde{x}}_n\big) =\\
    &\sum_{m=1}^{M} \sum_{n=1}^{N} \tr\big(\mvec{R}^\transpose (\mvec{\gamma}_2\mvec{\gamma}_2^\transpose - \mvec{\gamma}_1\mvec{\gamma}_1^\transpose)\mvec{R}\tau_{mn}\mvec{\tilde{x}}_n\mvec{\tilde{x}}_n^\transpose\big) = \\
    &\tr\bigg(\mvec{R}^\transpose (\mvec{\gamma}_2\mvec{\gamma}_2^\transpose - \mvec{\gamma}_1\mvec{\gamma}_1^\transpose)\mvec{R}\sum_{m=1}^{M} \sum_{n=1}^{N}\tau_{mn}\mvec{\tilde{x}}_n\mvec{\tilde{x}}_n^\transpose\bigg).
\end{align*}
By substituting the following matrices  
\begin{align*}
    &\mvec{A} = \mvec{\gamma}_2\mvec{\gamma}_2^\transpose-\mvec{\gamma}_1\mvec{\gamma}_1^\transpose \nonumber \\
    &\mvec{B}=\sum_{m=1}^{M} \sum_{n=1}^{N} \tau_{mn} \mvec{\tilde{x}}_n \mvec{\tilde{x}}_n^\transpose \\
    &\mvec{C} = \sum_{m=1}^{M} \sum_{n=1}^{N} \tau_{mn} \mvec{\tilde{x}}_n \mvec{\tilde{y}}_m^\transpose,
\end{align*}
and rearranging the terms, we can rewrite \eqref{eq:argminR} in a compact matrix form as
\begin{align}\label{solve_for_R}
    \mvec{R}^* = \argmin_{\mvec{R}\in SO(3)}\ 
    \beta\tr(\mvec{R}^\transpose\mvec{A}\mvec{R}\mvec{B})
    -\kappa \tr(\mvec{R} \mvec{C}).
\end{align}
Since the constraint in \eqref{eq:constrained} is the set of \textit{special orthogonal group} or the set of rotation matrices which are smooth, it is convenient to use Riemannian manifold optimization. We use the manifold optimization solver from the package in \cite{manopt}.

\subsubsection{Distribution Parameters Update:}\label{moment_estimate}
After updating the rotation matrix, we need to update distribution parameters as well. Recalling equation \eqref{eq:Q_expand}, we just retain terms that depend on the parameters and also substitute the optimal rotation matrix from previous step,
\begin{align}\label{eq:argmin_params}
    &(\kappa^*, \beta^*, \mvec{\gamma}_1^*, \mvec{\gamma}_2^*)
    = \argmin_{\mvec{\gamma}_1^\transpose\cdot\mvec{\gamma}_2=0}\ \sum_{m,n=1}^{M,N}
    \tau_{mn} \bigg(-\log c(\kappa, \beta)^{-1}\nonumber \\
    &-\kappa\mvec{\tilde{y}}_m^\transpose\mvec{R}^*\mvec{\tilde{x}}_n
    + \beta \big[(\mvec{\gamma}_2^\transpose\mvec{R}^*\mvec{\tilde{x}}_n)^2-
     (\mvec{\gamma}_1^\transpose\mvec{R}^*\mvec{\tilde{x}}_n)^2\big]\bigg).
\end{align}
There are some attempts to solve variant of this problem (in the context of mixture of Kent distributions) iteratively using BFGS quasi-Newton method with reparametrization of variables (\cite{GIMLOP}) or BSLM approach (\cite{BSLM}), but there is no guarantee to find a solution considering the nonconvex nature of the problem. A more convenient method to solve for parameters, as described in \cite{kent_mixture}, is the \textit{moment estimation} method. The method should be adapted for the mixture of Kent distribution, as it is only described before for the unimodal Kent distributions. Here we compute the \textit{weighted sample moments} as follows,
\begin{align*}
    \mvec{\bar{\tilde{x}}} &= \frac{\sum_{n=1}^{N} \Tau_n (\mvec{R}^*\mvec{\tilde{x}}_n)}{N_p} \\
    \mvec{S} &= \frac{\sum_{n=1}^{N} \Tau_n (\mvec{R}^*\mvec{\tilde{x}}_n)(\mvec{R}^*\mvec{\tilde{x}}_n)^\transpose}{N_p}
\end{align*}
where $\Tau_n = \sum_{m=1}^{M}\tau_{mn}$ and $N_p = \sum_{m=1}^{M} \sum_{n=1}^{N} \tau_{mn}$. Using the above \textit{weighted moments}, the rest of the method has been explained in \cite{Kent1982}, and we exclude the details here for brevity.

Using the moment estimate instead of ML estimate makes the EM algorithm to lose the theoretical convergence guarantees, however as stated in \cite{Kent1982}, the moment estimation is very close to ML estimate in case of small eccentricity $\frac{2\beta}{\kappa}$ or large $\kappa$, which is a usual case in practice.
\subsubsection{Rotation averaging and Translation:}
Finally we need to compute average of the rotations from different clusters and estimate the translation. Rotation averaging in Euclidean sense can be computed as,
\begin{align}\label{eq:rot_avg}
    \mvec{R} = \overline{\mvec{R}}\mvec{U}
    \text{diag}(\frac{1}{\sqrt{\Lambda_1}}, \frac{1}{\sqrt{\Lambda_2}}, \frac{s}{\sqrt{\Lambda_3}})\mvec{U}^\transpose
\end{align}
where $\overline{\mvec{R}} = \frac{\sum_{i}^{N} \mvec{R}_i^*}{N}$, $N^2\mvec{U}^\transpose\mvec{D}\mvec{U} = N^2\overline{\mvec{R}}^\transpose\overline{\mvec{R}}$, and $\mvec{D} = \text{diag}(\Lambda_1, \Lambda_2, \Lambda_3)$. Also $s=1$ if $\det(\overline{\mvec{R}})$ is positive and $s=-1$ otherwise (\cite{rot_avg}). $\mvec{R}_i^*$ are the optimal rotation matrices found by the algorithm for each cluster in the dataset. The translation vector also can easily be computed based on the positional mean of rotated \emph{observed} data and \emph{model} data, 
\begin{align}\label{eq:translatino_eq}
    \mvec{t}^* = \mvec{\mu}_y - \mvec{R}^* \mvec{\mu}_x.
\end{align}
This is valid, since it is the optimal translation in the sense of Euclidean norm which has been used in ICP-based methods as well. 

\section{Experimental Results}
\label{Expriment}
\begin{figure}[!t]
    \centering
    \includegraphics[width=0.475\linewidth]{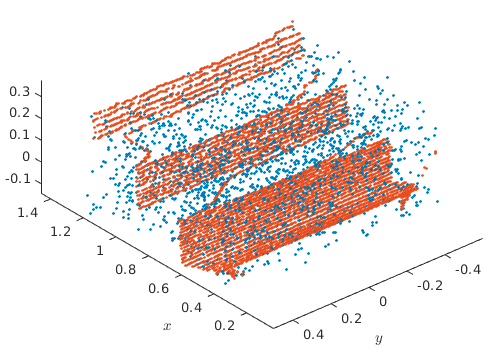}
    \includegraphics[width=0.475\linewidth]{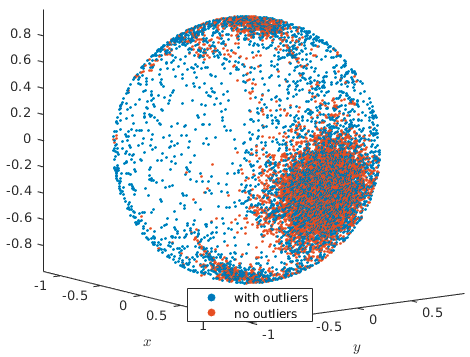}
    \includegraphics[width=0.475\linewidth]{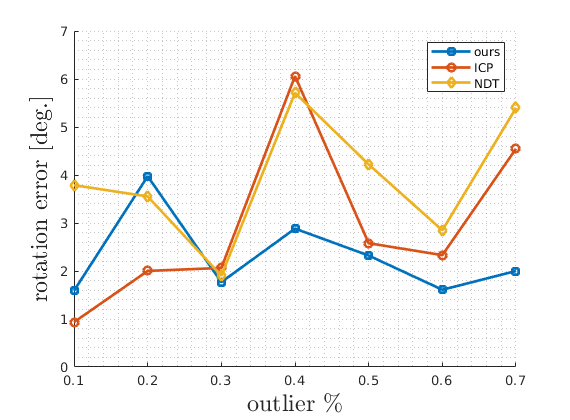}
    \includegraphics[width=0.475\linewidth]{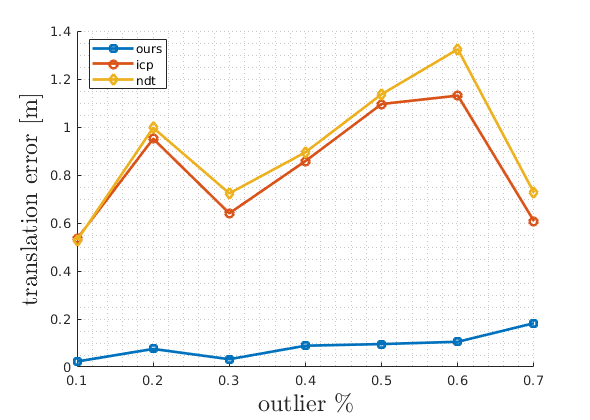}
    \caption{Top: Example of staircase pointcloud with $20\%$ outliers (blue points) and it's associated surface normals. Bottom: Comparison of rotation and translation errors.}
    \label{fig:outliers_comp}
\end{figure}
In this section the proposed method is tested and verified via 3D pointcloud registration on real indoor scanner pointclouds (\cite{ETH_Hauptgebaude}). We pre-processed the pointcloud frames before applying the algorithm. In the first stage, frames were down sampled to $10\%$ of the original size (about $10,000$ points) which was empirically chosen. In addition, all the processes on the surface normals described in \secref{surf_nromals_sub_sec} was performed. The proposed method was implemented in MATLAB. The algorithm was initialized with identity rotation matrix and zero translation vector. The distribution parameters was initialized in \emph{E}-step based on the current frame. We empirically found best results with $15$ neighborhood points for surface normals estimation. We adopt the error metric for comparison of estimated and true rotation matrices from \cite{Di_paper},
\begin{align*}
    e_{\mvec{\widehat{R}}} = \cos^{-1}\bigg(\frac{\tr(\mvec{R}^\transpose\mvec{\widehat{R}})-1}{2}\bigg)
\end{align*}
where $\mvec{\widehat{R}}$ and $\mvec{R}$ are estimated and true rotation matrix, respectively. And for the translation error we have $l2$-norm of the difference of estimated and true translation $e_{\mvec{\widehat{t}}} = ||\mvec{\widehat{t}} - \mvec{t}||_2$.
\subsection{Outlier Robustness}
We performed a set of point registration experiments comparing the robustness of the proposed method comparing with ICP and NDT methods. \figref{fig:outliers_comp} shows an example of frame with outlier points in blue. It can be seen that how the positional outliers impact and spread their associated surface normals on the unit sphere. The bottom plots illustrate the average of rotation and translation errors with different level of outliers injected into the pointclouds over $N_{trials}=100$ trials. As seen, ICP and NDT methods have inconsistent and higher error with regard to outliers which makes them unreliable in practice. In contrast, the proposed method has relatively lower errors and consistent with the level of outliers.
\subsection{Performance Evaluation}
\begin{figure}[!t]
    \centering
    \includegraphics[width=\linewidth]{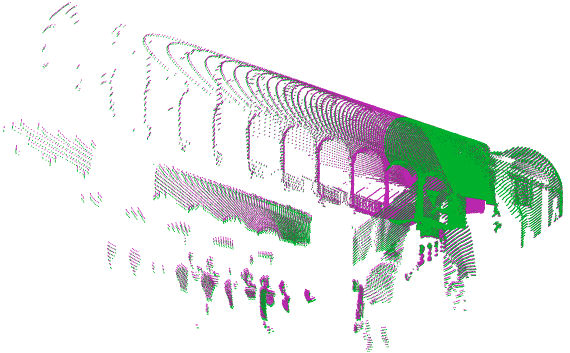}
    \caption{An example of matching quality (i.e. frames alignment) in two consecutive frames from indoor scanning data set \cite{ETH_Hauptgebaude} using the proposed method. Purple: the model pointcloud. Green: the transformed pointcloud.}
    \label{fig:examp.}
\end{figure}

The results from comparison of different point registration methods are tabulated in table~\ref{tb:results}. The results show the average rotation and translation error on the sequence of same frames from the dataset. For NDT (\cite{ndt}), the grid size is set as $2.0, 4.0, 2.0, 1.0$ meters with aligns the pointclouds in a finer-coarser-to-finer manner, and for CICP (\cite{CICP}) the scale factor is $\sigma= 1.0$ meter with a decay rate of 0.96. \figref{fig:examp.} represents an example of the point matching of two consecutive frames. The green is the pointcloud that is transformed back into the model pointcloud based on the estimated rigid transformation found by the proposed method.  
\begin{table}
    \begin{center}
        \caption{Comparison of PR Methods}\label{tb:results}
        \begin{tabular}{l|llllll}
            Trials & ICP & ICPp2pl & CICP & GICP & NDT & Ours \\
            \hline
            Avg. $e_{\mvec{\widehat{R}}}$ &1.27  & 0.57& 1.162& 0.23&  0.52&  0.66\\
            \hline
            Avg. $e_{\mvec{\widehat{t}}}$ &0.403 & 0.164& 0.260& 0.025& 0.070& 0.018\\ 
            \hline
        \end{tabular}
    \end{center}
\end{table}

As seen in the table~\ref{tb:results}, the proposed algorithm outperforms other methods with regard to translation error and generates satisfying rotation errors compared with the state-of-the-art NDT and GICP. It is reasonable to infer that the more information involved, the more accurate of the point registration algorithm. For instance, the ICP and CICP based on point-to-point distance produce the worse matching results. Likewise, GICP and NDT utlize more geometrical information like curvatures and more abstracted covariance matrices, which guarantees the better results. 

The proposed feature-based point registration method has two advantages over other methods: firstly only the directional information is utilized for rotation estimation and secondly the computationally correspondence phase in ICP- or NDT-based methods is avoided, but the registration results are still competitive even with NDT or GICP. On the other hand, since the indoor pointcloud is free of most of missing data comparing with outdoor scenes, we expect our method would perform better in the presence of outliers, occluded, and missing data as presented in the previous experiment case. 

Finally as stated in \secref{algorithm}, although using \emph{moment estimate} method makes the EM algorithm without convergence guarantee, since the algorithm utilizes the highly concentrated surface normals on the unit sphere (i.e. large $\kappa$) the moment estimate is closely related to ML estimates. In fact, the empirical results from our experiment show that \eqref{eq:Qfunc} the $Q$-function is not decreasing after each iteration of the algorithm.





\section{Conclusion}
\label{conclusion}
In this paper, our proposed feature-based method for 3D point registration was presented and evaluated with examples of indoor pointclouds. The algorithm incorporates surface normals in the scene as directional information to be used within our cohesive and probabilistic framework which improves the registration accuracy comparing to ICP-based methods. The method utilizes the fact that the translation-invariant property of surface normals decouples the estimation of rotation from translation. Additionally, the proposed method provides a robust mechanism that rejects outliers in the correspondence phase in a probabilistic fashion. The computation time comparison is not analyzed in this paper and one possible future direction can be reducing the computation time of the proposed algorithm.
\bibliography{ifacconf}

\begin{thebibliography}{29}
\providecommand{\natexlab}[1]{#1}
\providecommand{\url}[1]{\texttt{#1}}
\providecommand{\urlprefix}{URL }
\expandafter\ifx\csname urlstyle\endcsname\relax
  \providecommand{\doi}[1]{doi:\discretionary{}{}{}#1}\else
  \providecommand{\doi}{doi:\discretionary{}{}{}\begingroup
  \urlstyle{rm}\Url}\fi

\bibitem[{Banerjee et~al.(2005)Banerjee, Dhillon, Ghosh, and
  Sra}]{k_means_von_mises}
Banerjee, A., Dhillon, I.S., Ghosh, J., and Sra, S. (2005).
\newblock Clustering on the unit hypersphere using {V}on {M}ises-{F}isher
  distributions.
\newblock \emph{J. Mach. Learn. Res.}, 6, 1345--1382.

\bibitem[{Besl and McKay(1992)}]{ICP}
Besl, P.J. and McKay, N.D. (1992).
\newblock A method for registration of 3-{D} shapes.
\newblock \emph{IEEE Trans. Pattern Anal. Mach. Intell.}, 14(2), 239--256.

\bibitem[{Billings and Taylor(2015)}]{GIMLOP}
Billings, S. and Taylor, R. (2015).
\newblock Generalized iterative most likely oriented-point ({G}-{IMLOP})
  registration.
\newblock \emph{International Journal of Computer Assisted Radiology and
  Surgery}, 10(8), 1213--1226.

\bibitem[{Boumal et~al.(2014)Boumal, Mishra, Absil, and Sepulchre}]{manopt}
Boumal, N., Mishra, B., Absil, P.A., and Sepulchre, R. (2014).
\newblock {M}anopt, a {M}atlab toolbox for optimization on manifolds.
\newblock \emph{Journal of Machine Learning Research}, 15, 1455--1459.

\bibitem[{Du et~al.(2018)Du, Xu, Zhang, Zhang, Gao, and Chen}]{CICP}
Du, S., Xu, G., Zhang, S., Zhang, X., Gao, Y., and Chen, B. (2018).
\newblock Robust rigid registration algorithm based on pointwise correspondence
  and correntropy.
\newblock \emph{Pattern Recognition Letters}.

\bibitem[{Evangelidis and Horaud(2018)}]{multiple_GMM}
Evangelidis, G.D. and Horaud, R. (2018).
\newblock Joint alignment of multiple point sets with batch and incremental
  expectation-maximization.
\newblock \emph{{IEEE} Trans. Pattern Anal. Mach. Intell.}, 40(6), 1397--1410.

\bibitem[{Geiger et~al.(2013)Geiger, Lenz, Stiller, and Urtasun}]{KITTI}
Geiger, A., Lenz, P., Stiller, C., and Urtasun, R. (2013).
\newblock Vision meets robotics: The {KITTI} dataset.
\newblock \emph{International Journal of Robotics Research (IJRR)}.

\bibitem[{{Horaud} et~al.(2011){Horaud}, {Forbes}, {Yguel}, {Dewaele}, and
  {Zhang}}]{rigid_gmm}
{Horaud}, R., {Forbes}, F., {Yguel}, M., {Dewaele}, G., and {Zhang}, J. (2011).
\newblock Rigid and articulated point registration with expectation conditional
  maximization.
\newblock \emph{IEEE Transactions on Pattern Analysis and Machine
  Intelligence}, 33(3), 587--602.

\bibitem[{Hornik et~al.(2012)Hornik, Feinerer, Kober, and
  Buchta}]{spherical_k_means}
Hornik, K., Feinerer, I., Kober, M., and Buchta, C. (2012).
\newblock Spherical k-means clustering.
\newblock \emph{Journal of Statistical Software}, 50, 1--22.

\bibitem[{Jian and Vemuri(2011)}]{robust_gmm}
Jian, B. and Vemuri, B. (2011).
\newblock Robust point set registration using {G}aussian mixture models.
\newblock \emph{Pattern Analysis and Machine Intelligence, IEEE Transactions
  on}, 33, 1633 -- 1645.

\bibitem[{Kent(1982)}]{Kent1982}
Kent, J.T. (1982).
\newblock The {F}isher-{B}ingham distribution on the sphere.
\newblock \emph{Journal of the Royal Statistical Society. Series B
  (Methodological)}, 44(1), 71--80.

\bibitem[{Klasing et~al.(2009)Klasing, Althoff, Wollherr, and
  Buss}]{Compare_surface_normals}
Klasing, K., Althoff, D., Wollherr, D., and Buss, M. (2009).
\newblock Comparison of surface normal estimation methods for range sensing
  applications.
\newblock \emph{Proceedings - IEEE International Conference on Robotics and
  Automation}, 3206--3211.

\bibitem[{Low(2004)}]{p_to_pl_ICP}
Low, K.L. (2004).
\newblock Linear least-squares optimization for point-to-plane {ICP} surface
  registration.

\bibitem[{Luo and Hancock(2001)}]{soft_assgin2}
Luo, B. and Hancock, E.R. (2001).
\newblock Structural graph matching using the {EM} algorithm and singular value
  decomposition.
\newblock \emph{IEEE Trans. Pattern Anal. Mach. Intell.}, 23(10), 1120--1136.

\bibitem[{{Min} et~al.(2019){Min}, {Wang}, and {Meng}}]{RG_register}
{Min}, Z., {Wang}, J., and {Meng}, M.Q. (2019).
\newblock Robust generalized point cloud registration with orientational data
  based on expectation maximization.
\newblock \emph{IEEE Transactions on Automation Science and Engineering},
  1--15.

\bibitem[{Moakher(2002)}]{rot_avg}
Moakher, M. (2002).
\newblock Means and averaging in the group of rotations.
\newblock \emph{SIAM J. Matrix Anal. Appl.}, 24(1), 1--16.

\bibitem[{Myronenko and Song(2010)}]{soft_assign3_GMM}
Myronenko, A. and Song, X. (2010).
\newblock Point set registration: Coherent point drift.
\newblock \emph{IEEE transactions on pattern analysis and machine
  intelligence}, 32, 2262--75.

\bibitem[{Nguyen(2017)}]{BSLM}
Nguyen, H. (2017).
\newblock A novel algorithm for clustering of data on the unit sphere via
  mixture models.

\bibitem[{Peel et~al.(2001)Peel, Whiten, and McLachlan}]{kent_mixture}
Peel, D., Whiten, W.J., and McLachlan, G.J. (2001).
\newblock Fitting mixtures of {K}ent distributions to aid in joint set
  identification.
\newblock \emph{Journal of the American Statistical Association}, 96(453),
  56--63.

\bibitem[{Pomerleau et~al.(2012)Pomerleau, Liu, Colas, and
  Siegwart}]{ETH_Hauptgebaude}
Pomerleau, F., Liu, M., Colas, F., and Siegwart, R. (2012).
\newblock {Challenging data sets for point cloud registration algorithms}.
\newblock \emph{The International Journal of Robotics Research}, 31(14),
  1705--1711.

\bibitem[{Rangarajan et~al.(1996)Rangarajan, Mjolsness, Pappu, Davachi,
  Goldman-Rakic, and Duncan}]{soft_assgin1}
Rangarajan, A., Mjolsness, E., Pappu, S., Davachi, L., Goldman-Rakic, P.S., and
  Duncan, J.S. (1996).
\newblock \emph{A robust point matching algorithm for autoradiograph
  alignment}, 277--286.
\newblock Springer Berlin Heidelberg, Berlin, Heidelberg.

\bibitem[{Rusu(2013)}]{consistent_normals}
Rusu, R.B. (2013).
\newblock \emph{Semantic 3D Object Maps for Everyday Robot Manipulation}.
\newblock Springer Publishing Company, Incorporated.

\bibitem[{Rusu et~al.(2008)Rusu, Marton, Blodow, Dolha, and Beetz}]{outlier}
Rusu, R.B., Marton, Z.C., Blodow, N., Dolha, M., and Beetz, M. (2008).
\newblock Towards 3{D} point cloud based object maps for household
  environments.
\newblock \emph{Robot. Auton. Syst.}, 56(11), 927--941.

\bibitem[{Segal et~al.(2009)Segal, Hähnel, and Thrun}]{GICP}
Segal, A., Hähnel, D., and Thrun, S. (2009).
\newblock Generalized-{ICP}.
\newblock \emph{Proc. of Robotics: Science and Systems}.

\bibitem[{Serafin and Grisetti(2015)}]{pl_to_pl_ICP}
Serafin, J. and Grisetti, G. (2015).
\newblock {NICP}: Dense normal based point cloud registration.
\newblock \emph{Proceedings of the ... IEEE/RSJ International Conference on
  Intelligent Robots and Systems. IEEE/RSJ International Conference on
  Intelligent Robots and Systems}, 742--749.

\bibitem[{{Stoyanov} et~al.(2012){Stoyanov}, {Magnusson}, and
  {Lilienthal}}]{ndt}
{Stoyanov}, T., {Magnusson}, M., and {Lilienthal}, A.J. (2012).
\newblock Point set registration through minimization of the l2 distance
  between 3d-ndt models.
\newblock In \emph{2012 IEEE International Conference on Robotics and
  Automation}, 5196--5201.

\bibitem[{Thomas et~al.(2019)Thomas, Wasenm{\"{u}}ller, and Stricker}]{Delio}
Thomas, Q.M., Wasenm{\"{u}}ller, O., and Stricker, D. (2019).
\newblock De{L}i{O}: Decoupled {L}i{DAR} odometry.
\newblock \emph{CoRR}, abs/1904.12667.

\bibitem[{Wang et~al.(2018)Wang, Xue, Tao, Zhong, Cui, Du, and
  Zheng}]{Di_paper}
Wang, D., Xue, J., Tao, Z., Zhong, Y., Cui, D., Du, S., and Zheng, N. (2018).
\newblock Accurate mix-norm-based scan matching.
\newblock In \emph{2018 IEEE/RSJ International Conference on Intelligent Robots
  and Systems (IROS)}, 1665--1671. IEEE.

\bibitem[{Zhang(1994)}]{outlier_removal}
Zhang, Z. (1994).
\newblock Iterative point matching for registration of free-form curves and
  surfaces.
\newblock \emph{International journal of computer vision}, 13, 119--152.

\end{thebibliography}

\end{document}